\documentclass[lettersize,journal]{IEEEtran}
\usepackage{amsmath,amsfonts}
\usepackage{algorithmic}
\usepackage{algorithm}
\usepackage{array}
\usepackage{textcomp}
\usepackage{stfloats}
\usepackage{url}
\usepackage{verbatim}
\usepackage{graphicx}
\usepackage{caption}
\usepackage{colortbl}
\usepackage{multirow}
\usepackage{multicol}
\usepackage{subcaption}
\usepackage{cite}
\usepackage[T1]{fontenc}  
\begin{document}

\title{Transmit Weights, Not Features: Orthogonal-Basis Aided Wireless Point-Cloud Transmission}
\newcommand{\coauthor}[1]{\textsuperscript{*}#1}
\author{
        {Junlin Chang},
        {Yubo Han}, 
        {Hang Yue},
        {John S Thompson},~\IEEEmembership{Fellow,~IEEE}
        {Rongke Liu},~\IEEEmembership{Senior Member,~IEEE}
        
\thanks {This work was supported in part by Shenzhen Fundamental Research Program under Grant JCYJ20220818103413029 and in part by Shenzhen Science and Technology Program under Grant GJHZ20220913144207013.}
\thanks{ Junlin Chang is with Beihang University and Pengcheng  Laboratory ( Changjunlin@buaa.edu.cn; changjl@pcl.ac.cn). Yubo Han and Hang Yue are with Beihang University (yubo.han@buaa.edu.cn; yue\_hang@buaa.edu.cn). Rongke Liu is with the Shenzhen Institute of Beihang University and Beihang University (rongke\_liu@buaa.edu.cn).}
\thanks{John S Thompson, Institute for Imaging, Data and Communications (IDCOM), School of Engineering, University of Edinburgh, Edinburgh, EH9 3BF, UK.(john.thompson@ed.ac.uk)}
\thanks{ Junlin Chang and Yubo Han make equal contributions. The corresponding author is Rongke Liu.(rongke\_liu@buaa.edu.cn)}
}

\markboth{Journal of \LaTeX\ Class Files,~Vol.~14, No.~8, August~2021}%
{Shell \MakeLowercase{\textit{et al.}}: A Sample Article Using IEEEtran.cls for IEEE Journals}


\maketitle

\begin{abstract}
The widespread adoption of depth sensors has substantially lowered the barrier to point-cloud acquisition. This letter proposes a semantic wireless transmission framework for three dimension (3D) point clouds built on Deep Joint Source–Channel Coding (DeepJSCC). Instead of sending raw features, the transmitter predicts combination weights over a receiver-side semantic orthogonal feature pool, enabling compact representations and robust reconstruction. A folding-based decoder deforms a 2D grid into 3D, enforcing manifold continuity while preserving geometric fidelity. Trained with Chamfer Distance (CD) and an orthogonality regularizer, the system is evaluated on ModelNet40 across varying Signal-to-Noise Ratios (SNRs) and bandwidths. Results show performance on par with SEmantic Point cloud Transmission (SEPT) at high bandwidth and clear gains in bandwidth-constrained regimes, with consistent improvements in both Peak Signal-to-Noise Ratio (PSNR) and CD. Ablation experiments confirm the benefits of orthogonalization and the folding prior.
\end{abstract}

\begin{IEEEkeywords}
Semantic communication, point cloud transmission, joint source-channel coding, deep learning.
\end{IEEEkeywords}

\section{Introduction}
\IEEEPARstart{W}{ith} the rapid proliferation of depth sensors such as LiDAR and RGB-D cameras, 3D point clouds, characterized by features including spatial coordinates, color attributes, and surface normals, have attracted increasing attention from the computer vision and computer graphics communities. Their applications span a wide range of domains such as autonomous driving~\cite{yang2024visual}, virtual/augmented reality (VR/AR)~\cite{casado2023rendering}, and robotics~\cite{wang2021trajectory}. In these contexts, wireless transmission plays a pivotal role in enabling the mobility, scalability, and accessibility of point cloud data. Conventional wireless transmission techniques, however, generally neglect the semantic information embedded in the source data during encoding. This leads to low transmission efficiency and strong vulnerability to noise perturbations, particularly in low signal-to-noise ratio (SNR) regimes where a severe cliff effect for bit errors may occur. Consequently, there is an urgent demand for efficient and robust point cloud transmission frameworks.

Semantic communication has recently emerged as a new paradigm for next-generation communication systems. By extracting and encoding semantic representations, semantic communication processes data in the semantic domain rather than the signal domain, thereby achieving high compression rates while retaining task-relevant information. Its core principle lies in capturing essential features and discarding redundant content, which directly enhances decoding performance and transmission reliability. In general, semantic communication systems leverage encoder–decoder architectures tailored to specific data modalities. Unlike such regular data formats (e.g., images with structured pixel arrangements), point clouds are unordered and irregular 3D point sets, which makes their semantic representation and wireless transmission particularly challenging. Effective transmission requires accurate modeling of geometric structures in three-dimensional space to ensure faithful recovery of spatial coordinates at the receiver.

Existing research on point cloud communication has primarily concentrated on the network layer of the Open System Interconnect (OSI) reference model, with relatively little attention devoted to physical-layer impairments such as channel fading and interference. For instance, AITransfer~\cite{zhu2022semantic} proposes an end-to-end framework for point cloud feature extraction and reconstruction, enhanced by deep reinforcement learning (DRL) to achieve adaptive control under dynamic network conditions. Similarly, ISCom ~\cite{huang2023iscom} adopts a two-stage region-of-interest (ROI) strategy that leverages user motion trajectory and saliency information, coupled with an online DRL-based scheduler to accommodate heterogeneous devices and time-varying network environments. Although these approaches enhance robustness and fidelity from a network-layer perspective, they neglect the effects of wireless channel impairments, thereby limiting their practicality in real-world wireless transmission scenarios. The most related method for our work is SEPT~\cite{bian2024wireless}, which employs a multi-scale encoder–decoder architecture: point clouds are iteratively downsampled at the transmitter and reconstructed at the receiver using latent variable recovery and offset-based upsampling. Although SEPT~\cite{bian2024wireless} demonstrates strong performance under moderate bandwidth constraints, its effectiveness significantly degrades in extremely limited bandwidth conditions, thereby restricting its applicability to such systems.

It is worth noting that natural data often exhibits approximate low-rank properties, which have been successfully exploited in diverse modalities such as images~\cite{gu2014weighted, han2022hyperspectral}, videos~\cite{ji2010robust}, audio~\cite{li2006singing}, and point clouds~\cite{chen2019multi, chen2025efficient}. Motivated by this property, we propose a novel wireless point cloud semantic transmission system built upon the Deep Joint Source–Channel Coding (DeepJSCC) framework~\cite{bourtsoulatze2019deep} to achieve efficient and reliable communication over wireless channels. Specifically, we construct a point cloud semantic orthogonal basis set (termed a feature pool) at the receiver. Instead of directly transmitting semantic features, the transmitter predicts the combined weights of these bases. The receiver then reconstructs semantic representations via weighted basis aggregation, followed by decoding the point cloud’s central point set. Finally, a folding-based decoder introduces an implicit 2D grid prior, which effectively addresses the irregularity of point cloud data while ensuring geometric fidelity. This design enables efficient semantic compression, robust wireless transmission, and high-quality reconstruction of point clouds even under severely constrained channel conditions.







The remainder of this paper is organized as follows. Sec.~\ref{sec:sys} presents the system model; Sec.~\ref{sec:method} details the proposed methodology; Sec.~\ref{sec:exp} provides simulation results and performance analysis; and Sec.~\ref{sec:con} summaries the paper and gives research directions.

\section{System Model}
\label{sec:sys}
In this section, we present the semantic communication framework for efficient point cloud transmission over wireless channels, together with the quantitative performance metrics used for evaluating transmission quality.

\subsection{Point Cloud Semantic Communication System}

We consider the transmission of 3D point clouds over an additive white Gaussian noise (AWGN) channel. A point cloud is represented as
\begin{equation}
P = \{\mathbf{p}_i\}_{i=1}^N, \quad \mathbf{p}_i = (\mathbf{x}_i, \mathbf{f}_i)
\end{equation}
where $N$ denotes the number of points. Each $\mathbf{x}_i \in \mathbb{R}^3$ corresponds to the 3D spatial coordinates, while $\mathbf{f}_i \in \mathbb{R}^C$ represents additional attributes such as normals or colors, with $C$ denoting the attribute dimension. In this work, we focus on the most fundamental and widely studied scenario, where only the 3D coordinate vectors $\mathbf{x}_i$ are transmitted, while the feature vectors $\mathbf{f}_i$ are not.

\begin{figure}[t]
    \centering
    \includegraphics[width=0.86\linewidth]{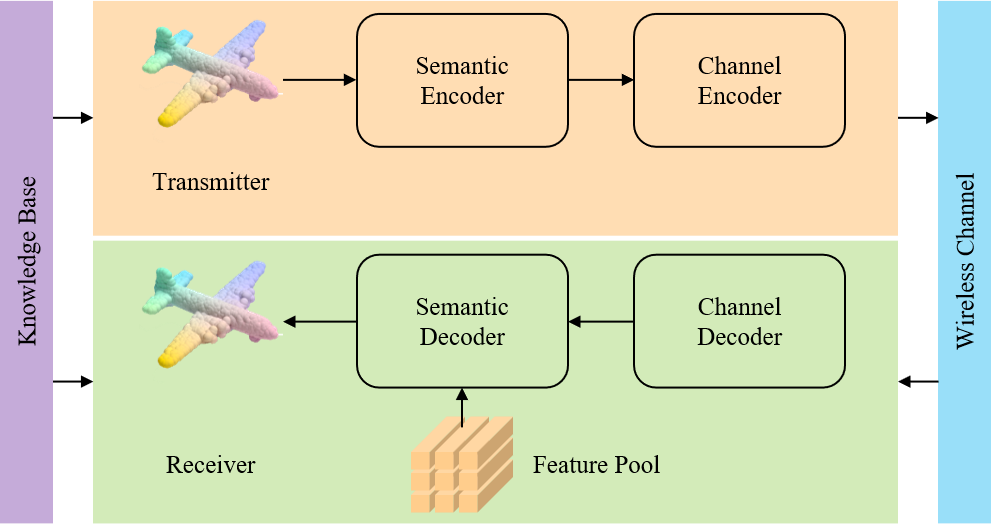}
    \caption{Overview of the proposed wireless point cloud semantic transmission system. 
    }
    \label{fig:architecture}
\end{figure}

The overall framework of the proposed semantic communication system is depicted in Fig.~\ref{fig:architecture}. The encoder and decoder in the DeepJSCC framework~\cite{bourtsoulatze2019deep} are denoted by $f_E(\cdot)$ and $f_D(\cdot)$, respectively. The encoder $f_E(\cdot)$ maps the input point cloud $P$ into a latent representation $\tilde{\mathbf{z}} \in \mathbb{R}^n$, where $n$ corresponds to the available channel bandwidth. Subsequently, power normalization is applied to the latent representation, producing a transmit codeword $\mathbf{z}$ with unit average power, which guarantees compliance with the transmission power constraint.
\begin{equation}
\mathbf{z} = \frac{\tilde{\mathbf{z}}}{\|\tilde{\mathbf{z}}\|_2}\label{eq:power}
\end{equation}
where $\|\mathbf{\tilde{\mathbf{z}}}\|_2 = \sqrt{\sum_i \tilde{z}_i^2}$ denotes the 
L2 norm of $\tilde{\mathbf{z}}$.

Before transmission, the real-valued codeword $\mathbf{z}$ is converted into a complex-valued vector $\hat{\mathbf{z}} \in \mathbb{C}^{n/2}$. The received signal at the receiver is then expressed as
\begin{equation}
\mathbf{y} = \hat{\mathbf{z}} + \mathbf{n}_g
\end{equation}
where $\mathbf{n}_g \sim \mathcal{N}(0, \sigma^2\mathbf{I})$ represents AWGN noise with unit variance, and $\sigma$ denotes the noise standard deviation determined by the signal-to-noise ratio (SNR). Finally, the decoder reconstructs the transmitted point cloud coordinates as
\begin{equation}
\hat{P} = f_D(\mathbf{y})
\end{equation}

\subsection{Evaluation Metrics for Point Cloud Transmission}

To comprehensively evaluate the quality of point cloud transmission, it is essential to adopt metrics that accurately measure both geometric fidelity and perceptual similarity between the reconstructed point cloud and the ground truth. In this work, we employ two peak signal-to-noise ratio (PSNR) variants~\cite{schwarz2018common}, namely D1 and D2, as well as the widely adopted Chamfer Distance (CD)~\cite{fan2017point}. These metrics have been extensively used in point cloud compression, registration, and reconstruction tasks. Let $A$ denote the reconstructed point cloud and $B$ the ground truth version of A.

\paragraph{D1 (Point-to-Point Error)}
The D1 metric evaluates the squared Euclidean distance between each point $\mathbf{a}_i$ in $A$ and its nearest neighbor $b_j$ in $B$:
\begin{equation}
e^{D1}_{A,B} = \frac{1}{N} \sum_{\mathbf{a}_i \in A} \min_{\mathbf{b}_j \in B} \|\mathbf{a}_i - \mathbf{b}_j\|_2^2
\end{equation}
This measure directly reflects the point-wise reconstruction error and is sensitive to small geometric misalignments.

\paragraph{D2 (Point-to-Surface Error)}
The D2 metric incorporates local surface information by considering normals:
\begin{equation}
e^{D2}_{A,B} = \frac{1}{N} \sum_{\mathbf{a}_i \in A} \big((\mathbf{a}_i - \mathbf{b}_j) \cdot \mathbf{n}_i\big)
\end{equation}
where $\mathbf{n}_i$ denotes the surface normal at $\mathbf{a}_i$ computed by~\cite{schwarz2018common}. This metric assesses the consistency between the reconstructed points and the underlying tangent plane of the original geometry, thereby capturing surface-level fidelity.

\paragraph{PSNR for D1 and D2}
Following~\cite{schwarz2018common}, the PSNR corresponding to D1 and D2 is defined as
\begin{equation}
\mathrm{PSNR} = 10 \log_{10}\!\left(\frac{3p^2}{\max(e_{A,B}^{D_i}, e_{B,A}^{D_i})}\right), i=1,2
\end{equation}
where $p$ represents the geometric resolution of the model. Since the input point clouds are normalized into the unit cube $[0,1]$, we set $p=1$.

\paragraph{CD}
The Chamfer Distance is a symmetric measure widely adopted in point cloud reconstruction and generation due to its robustness to noise and outliers:
\begin{equation}
\mathrm{CD}(A,B) = e^{D1}_{A,B} + e^{D1}_{B,A}
\label{eq:cd}
\end{equation}
Unlike D1 and D2, CD accounts for bidirectional nearest-neighbor consistency, making it more suitable for evaluating global geometric similarity and robustness to density variations.

In summary, D1 captures local point-wise accuracy, D2 evaluates surface consistency, and CD emphasizes global geometric similarity while being robust to outliers. By jointly considering these metrics, we provide a comprehensive assessment of point cloud transmission performance from complementary perspectives.

\section{Proposed Method}
\label{sec:method}

In this section, we present the details of our proposed wireless point cloud semantic transmission system. The system follows the DeepJSCC paradigm~\cite{bourtsoulatze2019deep} and is composed of an encoder at the transmitter and a decoder at the receiver, each containing semantic and channel modules.

\subsection{DeepJSCC Encoder}
The encoder consists of two key components: the semantic encoder and the channel encoder.

\subsubsection{Semantic Encoder}
The semantic encoder extracts instance-relevant representations from the input point cloud, enabling compact and efficient transmission. Existing approaches often rely on carefully designed encoders trained from scratch to learn task-specific semantic features. However, such a strategy suffers from two major limitations: (i) the absence of prior knowledge and suitable parameter initialization leads to slower convergence and substantially higher computational costs; and (ii) training from scratch on limited data increases the risk of overfitting or convergence to suboptimal local minima. 

To address these issues, we adopt Point-BERT~\cite{yu2022point}, a pre-trained point cloud model, as the semantic encoder to exploit its strong representation capability. Specifically, a Mini-PointNet module first embeds the raw points into tokenized features, followed by 12 Transformer blocks to capture global semantic dependencies across points. By fine-tuning Point-BERT~\cite{yu2022point}, we transfer pre-trained knowledge from large datasets to wireless transmission tasks to boost performance and robustness.

\subsubsection{Channel Encoder}
The channel encoder compresses semantic features into a bandwidth-constrained latent representation. 
Formally, the latent code is computed as follows.

\begin{equation}
    \tilde{\mathbf{z}} = \gamma([\mathbf{x}_{cls}, \mathcal{M}(\mathbf{x})])
\end{equation}
where $\mathbf{x}_{cls}$ and $\mathbf{x}$ are the class token and feature tokens from Point-BERT, $\mathcal{M}(\cdot)$ denotes a max pooling operation, $[\cdot, \cdot]$ denotes concatenation, and $\gamma(\cdot)$ is an MLP. 

Finally, the latent code $\tilde{\mathbf{z}}$ is normalized to satisfy the unit average power constraint, as defined in (\ref{eq:power}).

\begin{figure}[t]
    \centering
    \begin{subfigure}{0.24\textwidth}
        \centering
        \includegraphics[height=3.4cm]{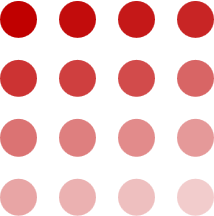}
        \caption{Grid points}
        \label{fig:grid}
    \end{subfigure}
    \hfill
    \begin{subfigure}{0.24\textwidth}
        \centering
        \includegraphics[height=3.4cm]{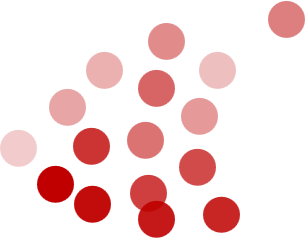}
        \caption{Deformed points}
        \label{fig:deformer}
    \end{subfigure}
    \caption{Illustration of the folding mechanism. A 2D grid prior is deformed into a 3D point cloud.}
    \label{fig:folding}
\end{figure}

\subsection{DeepJSCC Decoder}
At the receiver, the decoder is composed of a channel decoder and a semantic decoder.

\subsubsection{Channel Decoder}
The channel decoder restores semantic features from noisy signals transmitted over the AWGN channel. To enhance robustness, we employ a residual correction mechanism, where a two-layer MLP $\delta(\cdot)$ is applied iteratively to predict and cancel noise:

\begin{equation}
    \hat{\mathbf{h}} = \underbrace{\mathbf{y} - \delta(\mathbf{y})}_n
\end{equation}
where $n$ is the number of repetitions, which by default is set to 1.

This formulation transforms the recovery task into a noise estimation problem, analogous to denoising in diffusion-based models~\cite{ho2020denoising}. Such a design not only simplifies the reconstruction process but also improves robustness, since the network learns to explicitly suppress channel-induced perturbations rather than directly regressing complex semantic features.

\subsubsection{Semantic Decoder}
The semantic decoder reconstructs the 3D point cloud from the decoded semantic features. We construct a feature pool $\mathcal{O} = \{\mathbf{o}_i\}_{i=1}^N$, where $N$ is the number of feature vectors, serving as an orthogonal semantic basis. Applying a Softmax operation to $\hat{\mathbf{h}}$ yields weights $\boldsymbol{\alpha} = \{\alpha_i\}_{i=1}^N$, which are then aggregated as:

\begin{equation}
    \mathbf{F} = \sum_{i=1}^{N} \alpha_i \cdot \mathbf{o}_i
\end{equation}

Next, the center coordinates of the point cloud are predicted by a two-layer MLP $\rho(\cdot)$:

\begin{equation}
    \mathbf{C} = \rho(\mathbf{F})
\end{equation}

Finally, as illustrated in Fig.~\ref{fig:folding}, we adopt a folding-based decoder $\mathcal{D}(\cdot)$, which deforms a structured 2D grid prior $\mathbf{G}$ into the target 3D space, conditioned on both the aggregated semantic feature $\mathbf{F}$ and the predicted center coordinates $\mathbf{C}$:

\begin{equation}
    \hat{\mathbf{P}} = \mathbf{C} + \mathcal{D}([\mathbf{C}, \mathbf{F}, \mathbf{G}]).
\end{equation}

By introducing this implicit 2D manifold prior, the folding mechanism offers two main advantages: (i) it provides a strong spatial constraint that facilitates the reconstruction of smooth and continuous 3D surfaces, and (ii) it enables a compact decoding process, where complex 3D geometries are generated from a simple and low-dimensional grid through learnable deformation. This design significantly improves reconstruction fidelity while maintaining computational efficiency.

\subsection{Loss Function}
To ensure fair comparison with SEPT~\cite{bian2024wireless}, we employ CD as the primary reconstruction loss as defined in (\ref{eq:cd}). To encourage diversity and stability in the feature pool, we introduce an orthogonality regularization:
\begin{equation}
    \mathcal{L}_{ort} = \| \mathcal{O}\mathcal{O}^{T} - \mathbf{I} \|_{F}
\end{equation}
where $\mathbf{I}\in \mathbb{R}^{N \times N}$ is the identity matrix and $\|\cdot\|_{F}$ denotes the Frobenius norm. 

The final training objective is:
\begin{equation}\label{LOSS_function}
    \mathcal{L}_{all} = \mathrm{CD}(A,B) + \beta \mathcal{L}_{ort}
\end{equation}
where $\beta$ is a balancing factor, which defaults to 1.0.

\section{Experiment Results}
\label{sec:exp}

In this section, we present the experimental setup and results used to evaluate the reconstruction performance of our method.

\begin{figure*}[t]
    \centering
    \begin{subfigure}{0.31\textwidth}
        \centering
        \includegraphics[width=\textwidth]{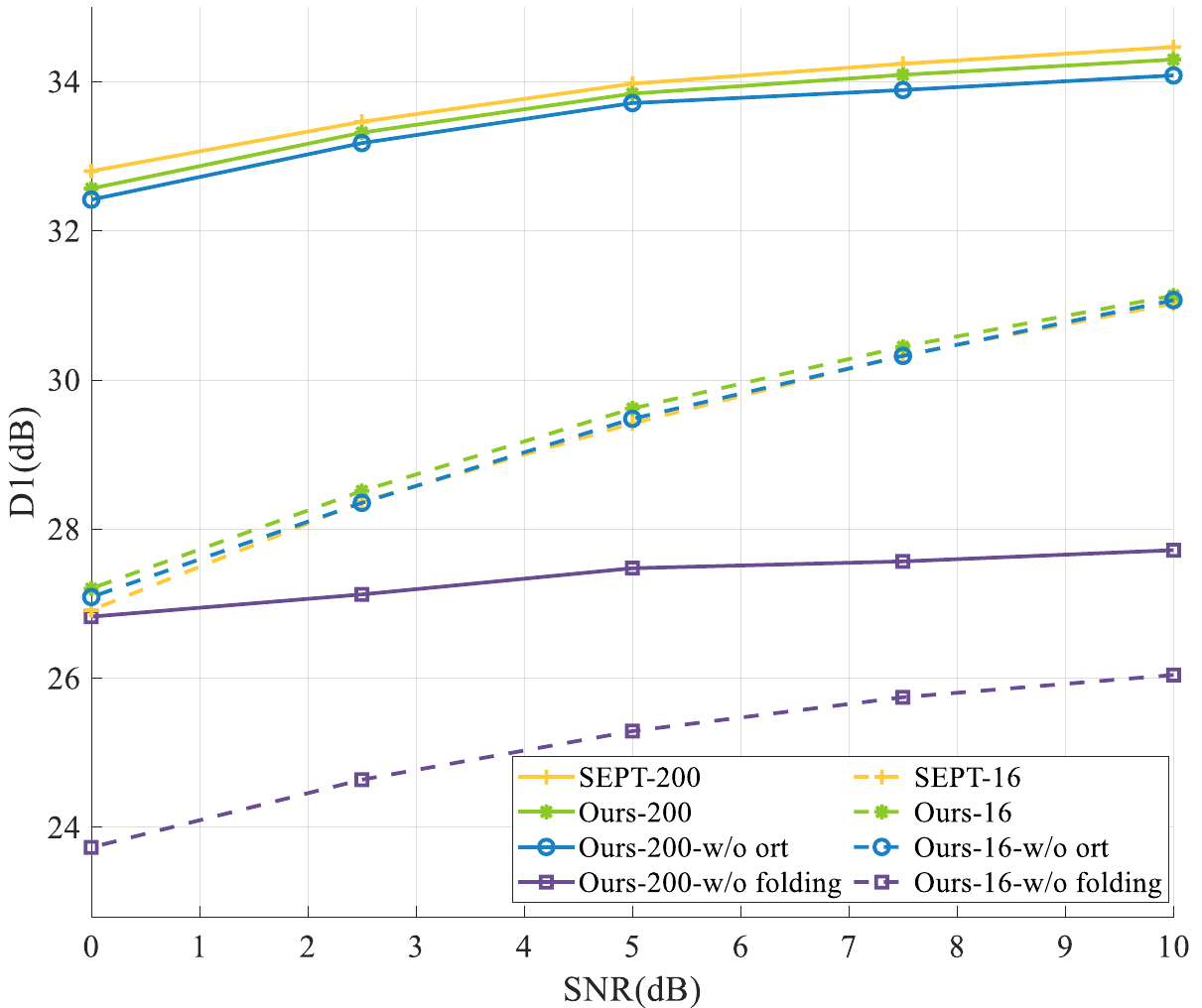}
        \caption{}
        \label{fig:d1}
    \end{subfigure}
    \hfill
    \begin{subfigure}{0.31\textwidth}
        \centering
        \includegraphics[width=\textwidth]{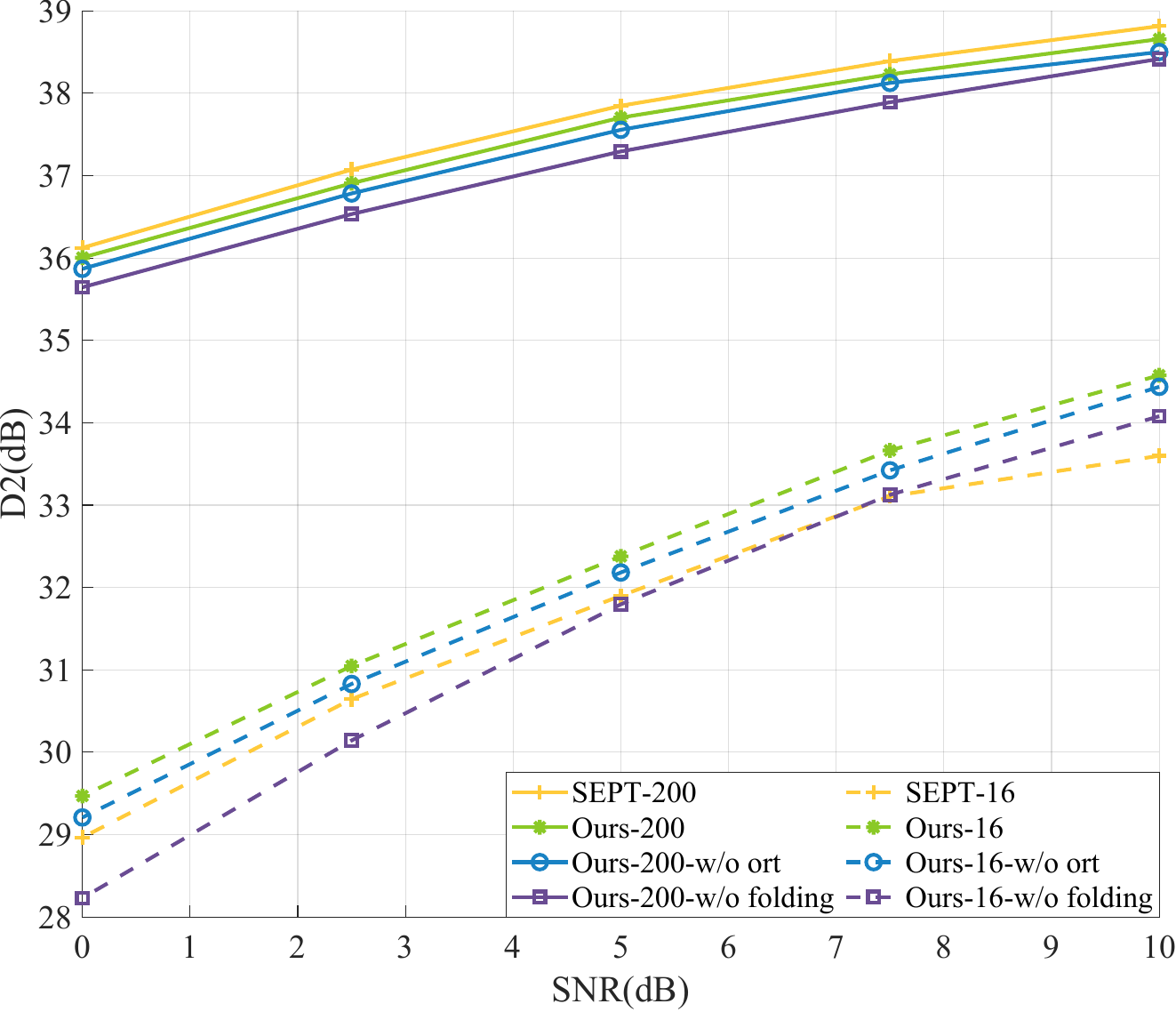}
        \caption{}
        \label{fig:d2}
    \end{subfigure}
    \hfill
    \begin{subfigure}{0.31\textwidth}
        \centering
        \includegraphics[width=\textwidth]{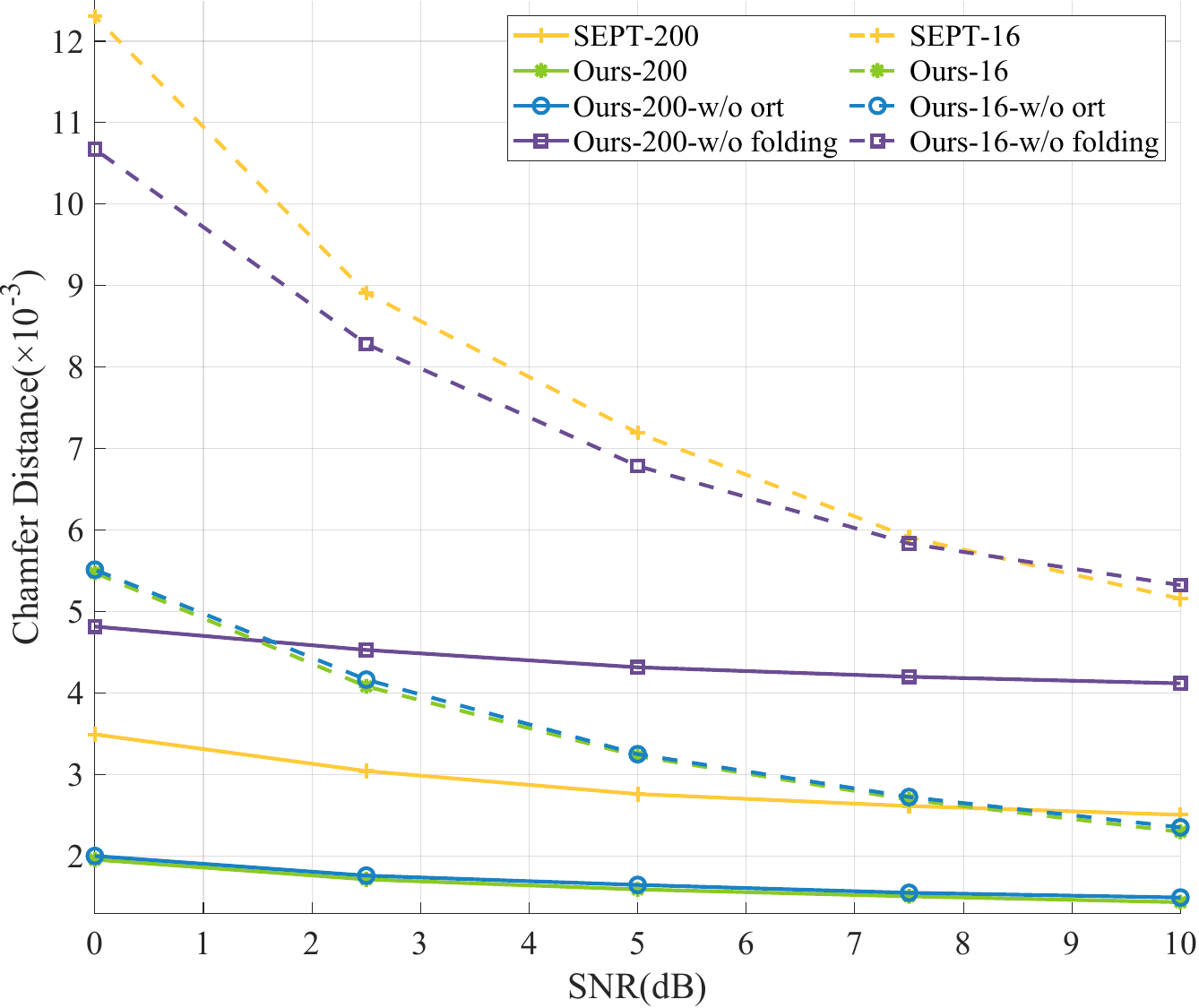}
        \caption{}
        \label{fig:cd}
    \end{subfigure}
    \caption{Comparison with SEPT~\cite{bian2024wireless} and ablation results. Subfigures (a)–(c) plot D1 (point-to-point), D2 (point-to-surface), and Chamfer Distance (CD) versus SNR. Legend entries of the form \emph{m–n} denote the method name (\emph{m}) and the available channel bandwidth (\emph{n}). \emph{(When D1/D2 are reported as PSNR, higher is better; for CD, lower is better.)}}
    \label{fig:main_results}
    \vspace{-10px}
\end{figure*}

\subsection{Experiment Setup}

We evaluate the proposed method on the ModelNet40 dataset~\cite{wu20153d}, which consists of 9,843 training samples and 2,468 test samples. Each point cloud is downsampled to 2,048 points using the farthest point sampling (FPS) algorithm~\cite{qi2017pointnet++}, and for fair comparison with SEPT~\cite{bian2024wireless}, all point clouds are normalized to the range $[-1,1]$. Without loss of generality, we communicate over an AWGN channel, consistent with the system model described in Sec.~\ref{sec:sys}. For the semantic encoder, we employ Point-BERT~\cite{yu2022point} pre-trained on ShapeNet~\cite{chang2015shapenet}, while the semantic decoder first predicts 128 central points, followed by a single folding operation to generate the complete point cloud with a grid size of $4 \times 4$, which achieves efficient decoding without sacrificing reconstruction accuracy. The model is trained using AdamW~\cite{Ilya2019decoupled} with an initial learning rate of $3 \times 10^{-4}$ under a cosine decay scheduler with 10 warm-up epochs, for a total of 200 epochs. Performance is evaluated under various SNRs using D1, D2, and CD metrics. All experiments are conducted on an Ubuntu 20.04 system equipped with an Intel Xeon Gold 6248R CPU @ 3.00GHz (96 cores) and four NVIDIA RTX A5000 GPUs, with a memory size of 128GB.

\begin{figure}[t]
    \centering
    \begin{subfigure}{0.22\textwidth}
        \centering
        \includegraphics[width=\textwidth]{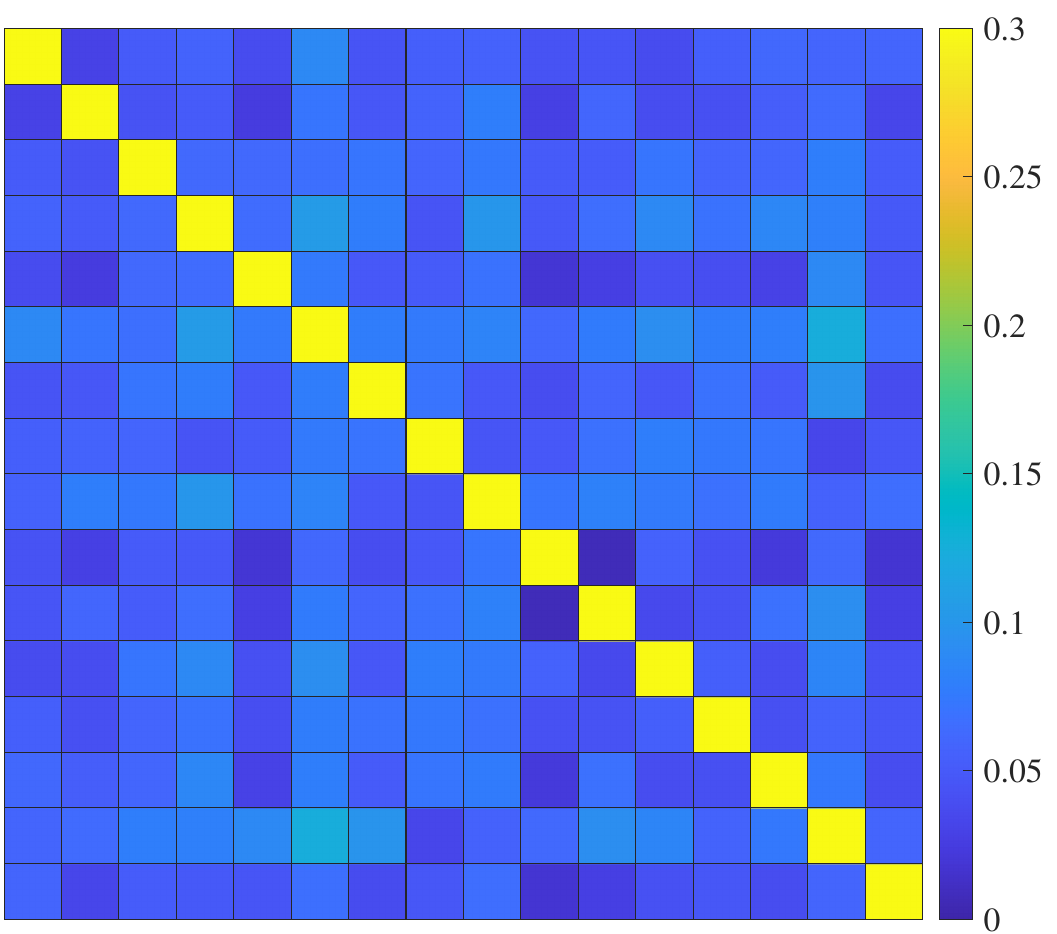}
        \caption{Ours-16}
        \label{fig:16}
    \end{subfigure}
    \hfill
    \begin{subfigure}{0.22\textwidth}
        \centering
        \includegraphics[width=\textwidth]{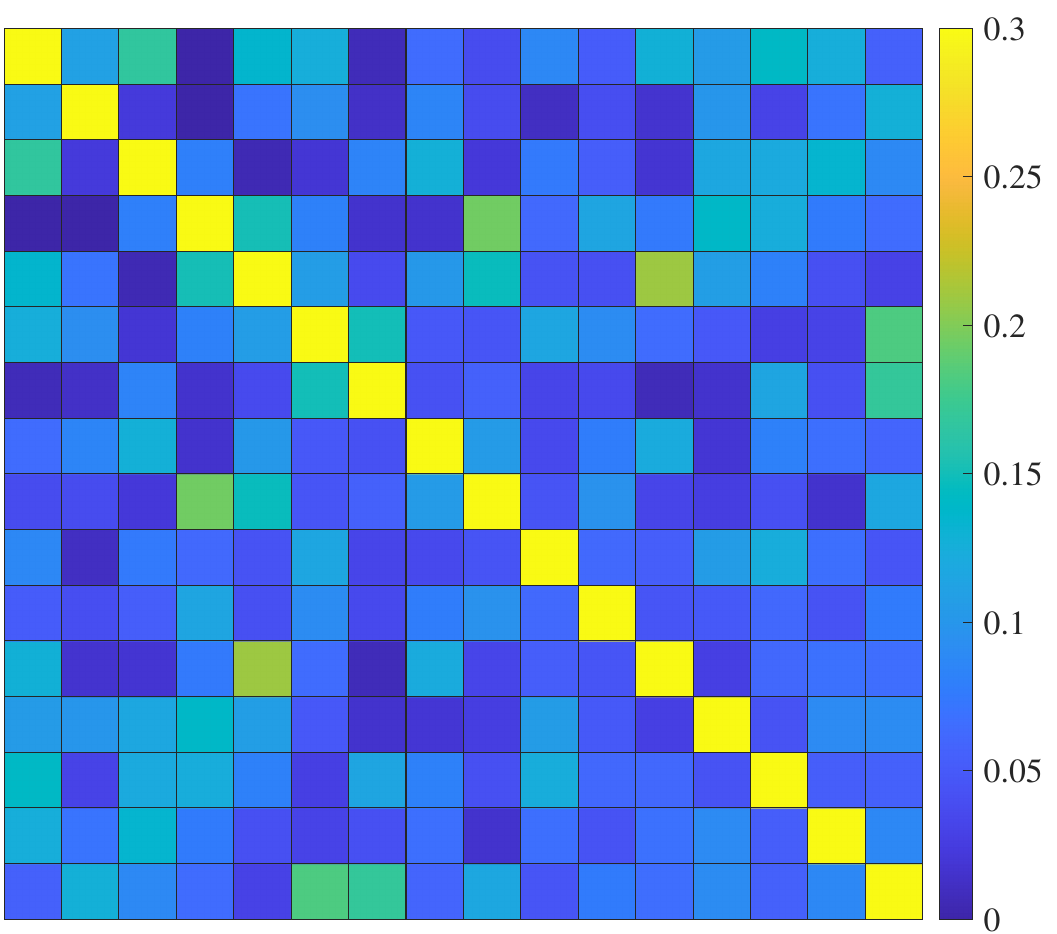}
        \caption{Ours-16-w/o ort}
        \label{fig:16_wo_ort}
    \end{subfigure}
    \hfill
    \begin{subfigure}{0.22\textwidth}
        \centering
        \includegraphics[width=\textwidth]{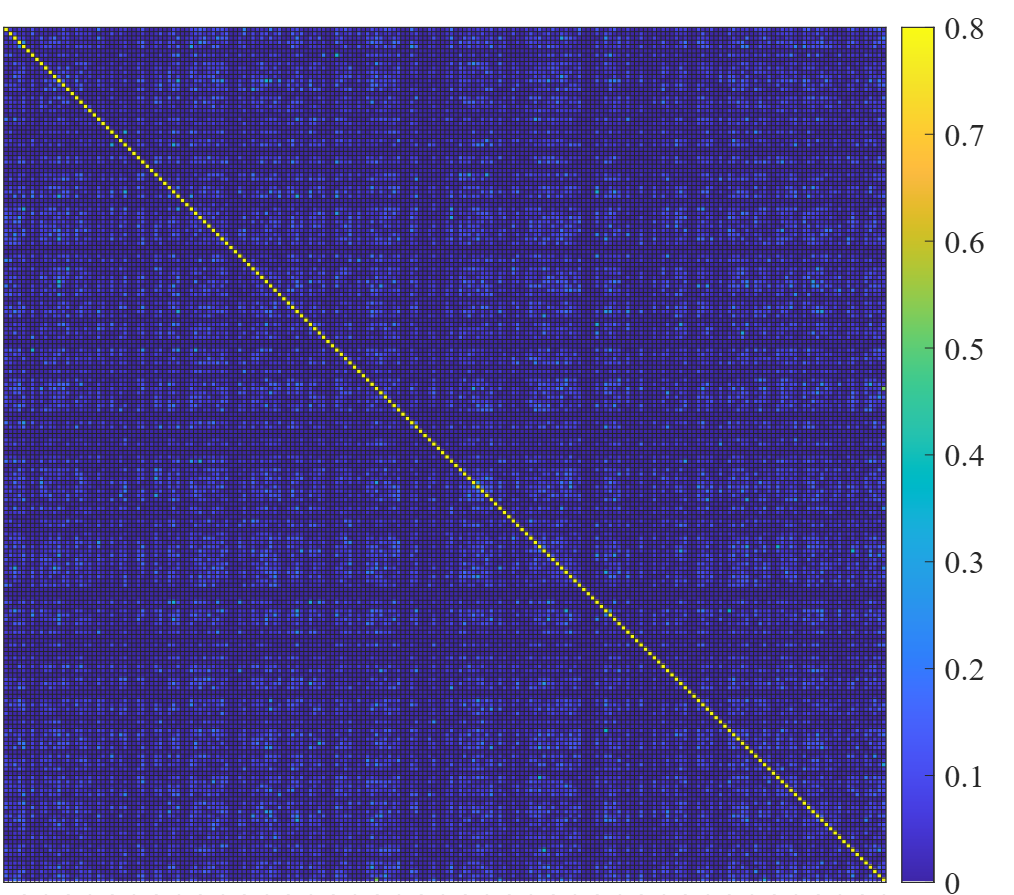}
        \caption{Ours-200}
        \label{fig:200}
    \end{subfigure}
    \hfill
    \begin{subfigure}{0.22\textwidth}
        \centering
        \includegraphics[width=\textwidth]{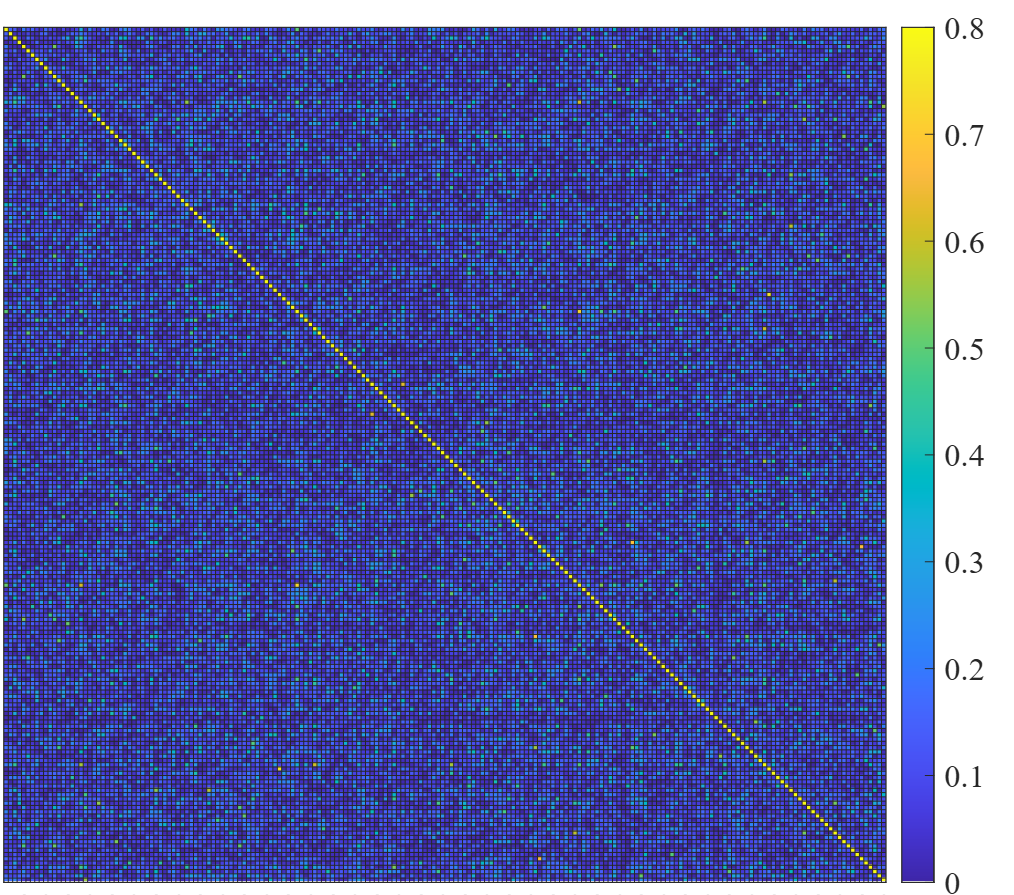}
        \caption{Ours-200-w/o ort}
        \label{fig:200_wo_ort}
    \end{subfigure}
    \caption{Visualization of feature pool orthogonalization. In order to better visualize orthogonality, all values are plotted in absolute magnitude and clipped to a fixed color scale.
    }
    \label{fig:ort_results}
    \vspace{-10px}
\end{figure}

\subsection{Reconstruction Performance}
We benchmark our approach against SEPT~\cite{bian2024wireless} under two channel bandwidth budgets, $n\in\{200,16\}$, over a range of SNRs. As shown in Fig.~\ref{fig:main_results}, Subfigures (a) to (c) summarize D1/D2 PSNR~\cite{schwarz2018common} and Chamfer Distance (CD)~\cite{fan2017point}. Under the high bandwidth setting ($n=200$) our method performs similarly to SEPT~\cite{bian2024wireless}. In bandwidth-constrained regimes ($n=16$) it consistently surpasses SEPT~\cite{bian2024wireless}, achieving higher D1/D2 and lower CD, which demonstrates robustness to channel impairments and efficient utilization of the available channel bandwidth. As expected, increasing bandwidth improves absolute performance for all methods, however, the degradation of our method when reducing $n$ from 200 to 16 is markedly smaller than that of SEPT~\cite{bian2024wireless}, indicating better use of limited bandwidth. Notably, our approach attains approximately half the CD of SEPT~\cite{bian2024wireless} across SNRs, and even at high SNR the $n=16$ variant achieves a lower CD than SEPT~\cite{bian2024wireless} with $n=200$, underscoring the effectiveness of the proposed encoding scheme.

\subsection{Ablation Study}
We conduct ablation studies to assess the contribution of two key components: (i) the orthogonality regularization on the feature pool and (ii) the folding–based decoder.

\begin{figure}[t]
    \centering
    \includegraphics[width=0.7\linewidth]{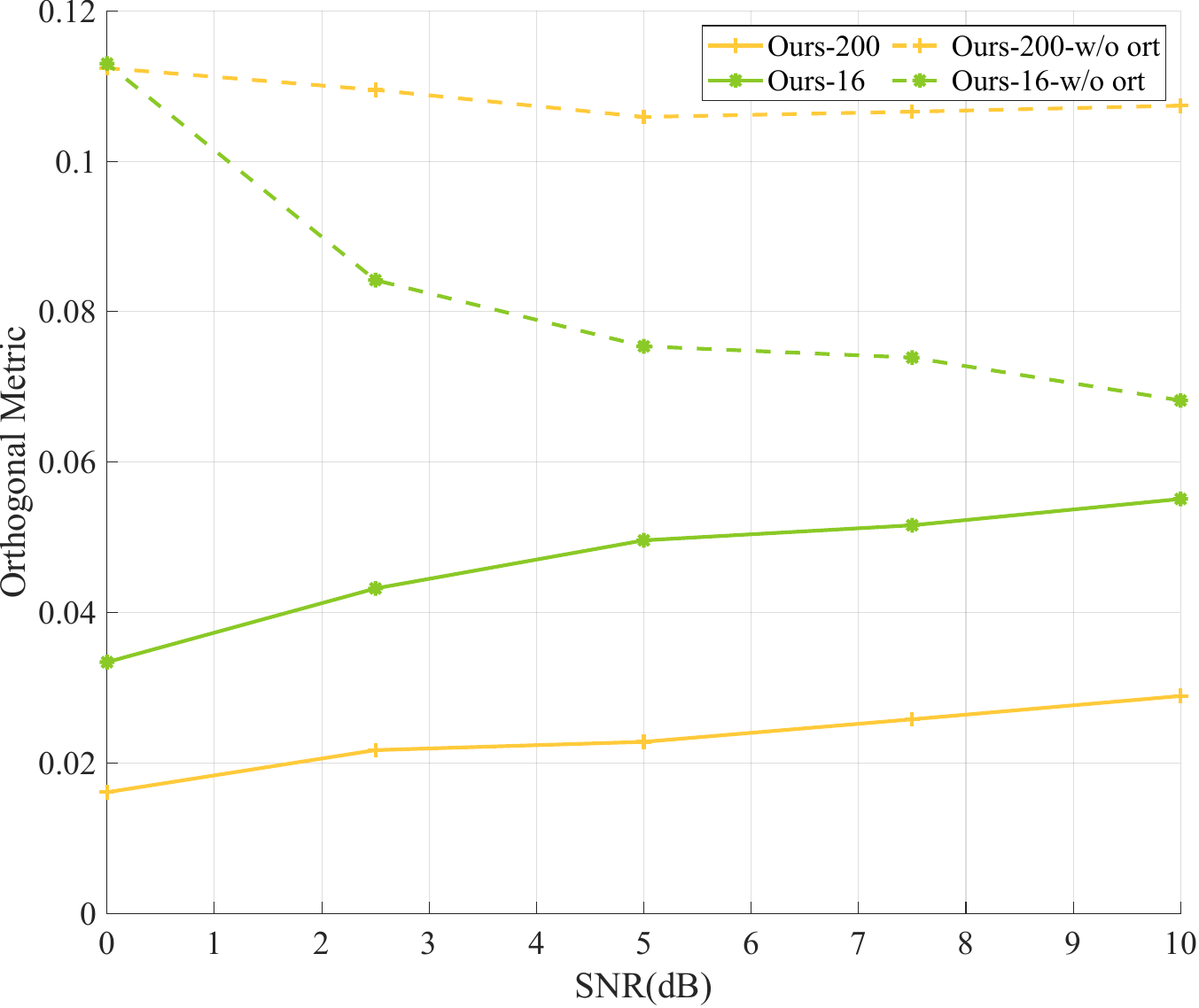}
    \caption{
    Orthogonal metrics under different SNRs.
    }
    \label{fig:ort_results1}
    \vspace{-10px}
\end{figure}

\paragraph{Effect of orthogonality regularization.}
Removing the orthogonality loss from \eqref{LOSS_function} - i.e. \textquotedbl w/o ort\textquotedbl \ leads to consistent degradation in D1/D2 and an increase in CD across SNRs in Fig.~\ref{fig:main_results}, showing that encouraging near–orthogonal bases reduces redundancy in the feature pool and stabilizes the weighting process. The visualizations in Fig.~\ref{fig:ort_results} further corroborate this effect: heatmaps of the Gram matrix $\mathcal{O}\mathcal{O}^{\top}$ exhibit a sharpened diagonal and suppressed off–diagonal entries after training with the orthogonality constraint, indicating improved basis decorrelation. Fig.~\ref{fig:ort_results1} plots the orthogonality metric $\|\mathcal{O}\mathcal{O}^{\top}-\mathbf{I}\|_{F}$ versus SNR. Models trained with the constraint maintain substantially lower values over all SNRs, confirming that the learned bases remain well–conditioned even under channel noise. 

\paragraph{Effect of the folding prior.}
Eliminating the folding module (w/o folding) produces the most pronounced performance drop in Fig.~\ref{fig:main_results}. These results confirm that the folding-based decoder improves reconstruction performance through a strong manifold prior that enforces surface continuity and global shape coherence. The benefit is especially pronounced in challenging cases where the decoder must recover plausible geometry from heavily compressed and noisy latent representations.


\section{Conclusion}
\label{sec:con}

This letter introduced a semantic joint source–channel coding framework for wireless 3D point-cloud transmission. The transmitter predicts weights over a receiver-side orthogonal feature pool, and a folding-based decoder enforces manifold continuity, yielding compact and channel-resilient representations. Trained with Chamfer Distance and an orthogonality regularizer, the method matches SEPT at high bandwidth and delivers consistent gains under bandwidth limited across SNRs. Ablations confirm the contributions of feature-pool orthogonalization and the folding mechanism. Current results are on AWGN and coordinate-only reconstruction and future work will extend to fading/MIMO channels and richer attributes.



{
    \small
    \bibliographystyle{IEEEtran}
    \bibliography{main}

@String(CVPR= {IEEE Conf. Comput. Vis. Pattern Recog.})

@String(ICLR = {Int. Conf. Learn. Represent.})

@String(CVPR  = {CVPR})

@String(ICLR  = {ICLR})

@inproceedings{yang2024visual,
  title={Visual point cloud forecasting enables scalable autonomous driving},
  author={Yang, Zetong and Chen, Li and Sun, Yanan and Li, Hongyang},
  booktitle={CVPR},
  pages={14673--14684},
  year={2024}
}

@article{casado2023rendering,
  title={Rendering massive indoor point clouds in virtual reality},
  author={Casado-Coscolla, Alvaro and Sanchez-Belenguer, Carlos and Wolfart, Erik and Sequeira, Vitor},
  journal={Virtual Reality},
  volume={27},
  number={3},
  pages={1859--1874},
  year={2023},
  publisher={Springer}
}

@article{wang2021trajectory,
  title={Trajectory planning and optimization for robotic machining based on measured point cloud},
  author={Wang, Gang and Li, Wenlong and Jiang, Cheng and Zhu, Dahu and Li, Zhongwei and Xu, Wei and Zhao, Huan and Ding, Han},
  journal={IEEE transactions on robotics},
  volume={38},
  number={3},
  pages={1621--1637},
  year={2021},
  publisher={IEEE}
}

@article{zhu2022semantic,
  title={A semantic-aware transmission with adaptive control scheme for volumetric video service},
  author={Zhu, Yuanwei and Huang, Yakun and Qiao, Xiuquan and Tan, Zhijie and Bai, Boyuan and Ma, Huadong and Dustdar, Schahram},
  journal={IEEE Transactions on Multimedia},
  volume={25},
  pages={7160--7172},
  year={2022},
  publisher={IEEE}
}

@article{huang2023iscom,
  title={ISCom: Interest-aware semantic communication scheme for point cloud video streaming on metaverse XR devices},
  author={Huang, Yakun and Bai, Boyuan and Zhu, Yuanwei and Qiao, Xiuquan and Su, Xiang and Yang, Lei and Zhang, Ping},
  journal={IEEE Journal on Selected Areas in Communications},
  volume={42},
  number={4},
  pages={1003--1021},
  year={2023},
  publisher={IEEE}
}

@inproceedings{bian2024wireless,
  title={Wireless point cloud transmission},
  author={Bian, Chenghong and Shao, Yulin and G{\"u}nd{\"u}z, Deniz},
  booktitle={SPAWC},
  pages={851--855},
  year={2024},
  organization={IEEE}
}

@article{bourtsoulatze2019deep,
  title={Deep joint source-channel coding for wireless image transmission},
  author={Bourtsoulatze, Eirina and Kurka, David Burth and G{\"u}nd{\"u}z, Deniz},
  journal={IEEE Transactions on Cognitive Communications and Networking},
  volume={5},
  number={3},
  pages={567--579},
  year={2019},
  publisher={IEEE}
}

@article{schwarz2018common,
  title={Common test conditions for point cloud compression},
  author={Schwarz, Sebastian and Martin-Cocher, Ga{\"e}lle and Flynn, David and Budagavi, Madhukar},
  journal={Document ISO/IEC JTC1/SC29/WG11 w17766, Ljubljana, Slovenia},
  year={2018}
}

@inproceedings{fan2017point,
  title={A point set generation network for 3d object reconstruction from a single image},
  author={Fan, Haoqiang and Su, Hao and Guibas, Leonidas J},
  booktitle={CVPR},
  pages={605--613},
  year={2017}
}

@inproceedings{yu2022point,
  title={Point-bert: Pre-training 3d point cloud transformers with masked point modeling},
  author={Yu, Xumin and Tang, Lulu and Rao, Yongming and Huang, Tiejun and Zhou, Jie and Lu, Jiwen},
  booktitle={CVPR},
  pages={19313--19322},
  year={2022}
}

@inproceedings{gu2014weighted,
  title={Weighted nuclear norm minimization with application to image denoising},
  author={Gu, Shuhang and Zhang, Lei and Zuo, Wangmeng and Feng, Xiangchu},
  booktitle={CVPR},
  pages={2862--2869},
  year={2014}
}

@inproceedings{ji2010robust,
  title={Robust video denoising using low rank matrix completion},
  author={Ji, Hui and Liu, Chaoqiang and Shen, Zuowei and Xu, Yuhong},
  booktitle={CVPR},
  pages={1791--1798},
  year={2010},
  organization={IEEE}
}

@inproceedings{li2006singing,
  title={Singing Voice Separation from Monaural Recordings.},
  author={Li, Yipeng and Wang, DeLiang},
  booktitle={ISMIR},
  volume={176},
  pages={179},
  year={2006}
}

@article{han2022hyperspectral,
  title={Hyperspectral and multispectral data fusion via nonlocal low-rank learning},
  author={Han, Yu and Bao, Wenxing and Zhang, Xiaowu and Ma, Xuan and Cao, Meng},
  journal={Journal of Applied Remote Sensing},
  volume={16},
  number={1},
  pages={016508--016508},
  year={2022},
  publisher={Society of Photo-Optical Instrumentation Engineers}
}

@article{chen2019multi,
  title={Multi-patch collaborative point cloud denoising via low-rank recovery with graph constraint},
  author={Chen, Honghua and Wei, Mingqiang and Sun, Yangxing and Xie, Xingyu and Wang, Jun},
  journal={IEEE transactions on visualization and computer graphics},
  volume={26},
  number={11},
  pages={3255--3270},
  year={2019},
  publisher={IEEE}
}

@article{chen2025efficient,
  title={Efficient Non-Local Point Cloud Denoising Using Curvature Entropy and $\gamma $-Norm Minimization},
  author={Chen, Jian and Gao, Feng and Chen, Pingping and Lin, Weisi},
  journal={IEEE Transactions on Visualization and Computer Graphics},
  year={2025},
  publisher={IEEE}
}

@article{ho2020denoising,
  title={Denoising diffusion probabilistic models},
  author={Ho, Jonathan and Jain, Ajay and Abbeel, Pieter},
  journal={NeurIPS},
  volume={33},
  pages={6840--6851},
  year={2020}
}

@inproceedings{wu20153d,
  title={3d shapenets: A deep representation for volumetric shapes},
  author={Wu, Zhirong and Song, Shuran and Khosla, Aditya and Yu, Fisher and Zhang, Linguang and Tang, Xiaoou and Xiao, Jianxiong},
  booktitle={CVPR},
  pages={1912--1920},
  year={2015}
}

@article{qi2017pointnet++,
  title={Pointnet++: Deep hierarchical feature learning on point sets in a metric space},
  author={Qi, Charles Ruizhongtai and Yi, Li and Su, Hao and Guibas, Leonidas J},
  journal={NeurIPS},
  volume={30},
  year={2017}
}

@article{chang2015shapenet,
  title={Shapenet: An information-rich 3d model repository},
  author={Chang, Angel X and Funkhouser, Thomas and Guibas, Leonidas and Hanrahan, Pat and Huang, Qixing and Li, Zimo and Savarese, Silvio and Savva, Manolis and Song, Shuran and Su, Hao and others},
  journal={arXiv preprint arXiv:1512.03012},
  year={2015}
}

@inproceedings{Ilya2019decoupled,
  title={Decoupled Weight Decay Regularization},
  author={Loshchilov, Ilya and Hutter, Frank},
  booktitle={ICLR},
  year={2019}
}
}

\vfill

\end{document}